\documentclass[conference]{IEEEtran}
\IEEEoverridecommandlockouts
\usepackage{cite}
\usepackage{amsmath,amssymb,amsfonts}
\usepackage{algorithmic}
\usepackage{graphicx}
\usepackage{textcomp}
\usepackage{xcolor}
\usepackage{cleveref}
\def\BibTeX{{\rm B\kern-.05em{\sc i\kern-.025em b}\kern-.08em
    T\kern-.1667em\lower.7ex\hbox{E}\kern-.125emX}}
\begin{document}

\title{The five \textit{I}s: Key principles for interpretable and safe conversational AI}

\author{
\IEEEauthorblockN{Mattias Wahde}
\IEEEauthorblockA{\textit{Department of Mechanics and Maritime Sciences} \\
\textit{Chalmers University of Technology}\\
Gothenburg, Sweden \\
mattias.wahde@chalmers.se}
\and
\IEEEauthorblockN{Marco Virgolin}
\IEEEauthorblockA{\textit{Department of Mechanics and Maritime Sciences} \\
\textit{Chalmers University of Technology}\\
Gothenburg, Sweden \\
marco.virgolin@chalmers.se}
}

\maketitle

\begin{abstract}
In this position paper, we present five key principles, namely \textit{interpretability}, \textit{inherent capability to explain}, \textit{independent data}, \textit{interactive learning}, and \textit{inquisitiveness}, for the development of conversational AI that, unlike the currently popular black box approaches, is transparent and accountable. At present, there is a growing concern with the use of black box statistical
language models: While displaying impressive average performance, such systems
are also prone to occasional spectacular failures, for which there is no clear remedy.
In an effort to initiate a discussion on possible alternatives, we outline and exemplify how our five principles enable the development of conversational AI systems that are transparent and thus safer for use. We also present some of the challenges inherent in the implementation of those principles.
\end{abstract}


\section{Introduction}
The field of conversational AI is in rapid development and it has important
applications in healthcare (for example elderly care), education, business
(for instance~customer service), and so on~\cite{LaranjoEtAl2018,BavarescoEtAl2020,WahdeVirgolin2021}. 
Those applications involve the use of conversational agents, which are systems 
intended for natural, multimodal interaction with human users, using text, speech, 
touch, gestures, and so on. 

At present, research in conversational AI is strongly dominated by black box 
approaches, in which deep neural networks (DNNs) are used to encode statistical 
language models that underlie the dialogue capabilities of a conversational agent. 
DNNs have achieved spectacular success in many AI subfields, 
notably image processing and speech processing~\cite{GoodfellowEtAl2016}. 
Thus, extending their use to conversational
AI may seem natural, and many promising results have indeed been presented
also in the case of conversational AI~\cite{OtterEtAl2020}.

However, there are considerable drawbacks with using black box
DNN-based systems, especially in high-stakes applications (such as, for example, healthcare)~\cite{Rudin2019,WahdeVirgolin2021}.
Due to their distributed nature, black box systems must generally
be trained with machine learning methods (deep learning) that require \emph{massive 
amounts of data}. Because the data sets are so large, it is nearly impossible to 
carry out a quality check to discard, e.g., unintended biases and sensitive content~\cite{BenderEtAl2021,BirhanePrabhu2021}.
Therefore, for conversational AI, the use of DNNs can result in conversational agents that, in addition
to being opaque, are unsafe for use, something that is discussed
in more detail in \Cref{sect:relatedwork} below.

While evidence on the unsuitability of data-hungry methods for safe conversational AI is mounting, a large part of current research focuses on 
creating methods to explain why black boxes make certain decisions, in an attempt to remedy the lack of transparency. 
However, explanation methods have their own   limitations~\cite{Rudin2019,Molnar2020}.

\begin{figure*}[ht]
  \centering
  \includegraphics[width=1.0\textwidth]{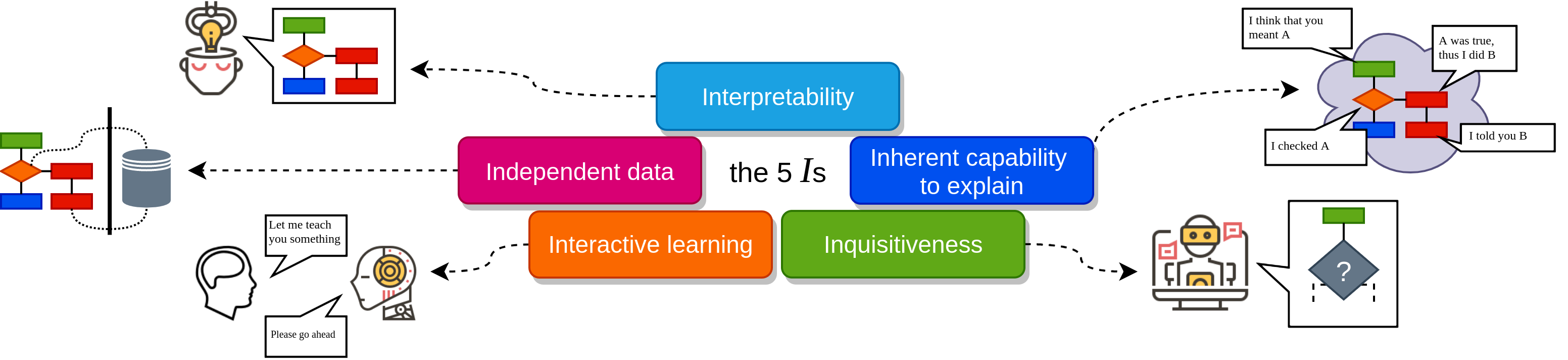}
  \caption{Five key principles for interpretable and safe conversational AI.}
  \label{fig:teaser}
\end{figure*}

To the best of our knowledge, no proposals exist yet on what could be radically different ways to move forward to create conversational AI that is transparent and accountable by construction.
In this position paper, we initiate this discussion by proposing \emph{five key design principles} that next-generation conversational AI should adhere to, in our view, to ensure transparency and accountability.
These are \emph{interpretability}, \emph{independent data}, \emph{inherent capability to explain}, \emph{interactive learning}, and \emph{inquisitiveness}\footnote{Here, we use the word \textit{inquisitiveness} in its positive sense, conveying
a meaning similar to the curiosity of a child.}; We call these the ``five Is'' of next-generation conversational AI; see Fig.~\ref{fig:teaser}.
In short, \emph{interpretability} and \emph{inherent capability to explain} are necessary aspects to ensure transparency and accountability while the conversational agent is being built and during use, respectively. 
\emph{Independent data} means that the agent should be easily adaptable to different application-specific knowledge bases, without the need of entirely re-working the way it handles information processing.
Last but not least, \emph{interactive learning} and \emph{inquisitiveness} represent our proposal 
to expand the arsenal of available training methods by incorporating a modality of
learning where the need for collecting and curating large amounts of data
is replaced by a process that is natural to humans.

The paper can be read as a kind of manifesto, emphasizing the five design principles listed above.

\section{Background \& Related Work}
\label{sect:relatedwork}
Black box systems such as DNNs have shown exceptional capability to capture and reproduce the statistical properties of language, at times generating documents and dialogues that are indistinguishable from those produced by humans~\cite{BrownEtAl2020,BartoliMedvet2020}.
However, there are several reasons for expressing concern with the use of black box
systems such as DNNs in conversational AI.

First of all, with DNNs, there is a near-complete opaqueness of the underlying decision-making required for the conversational agent to generate an utterance.
In fact, the decision-making of a DNN is the result of the concerted actions of myriads of neurons.
For example, the state-of-the-art DNN-based language model
GPT-3 uses 175 \textit{billion} parameters~\cite{BrownEtAl2020}. 
Even if such a system
can generate sensible answers to many queries, it is far from guaranteed to do so reliably~\cite{MarcusDavis2020}, and it is nearly impossible to foresee occasional spectacular failures~\cite{Daws2020}.
Furthermore, explanations of the underlying decision-making in such systems are generally not predictive in nature, but must rather be carried out 
on a case-by-case basis, ex post facto~\cite{Rudin2019,Molnar2020}.

Second, as briefly mentioned above, black box dialogue systems generally require huge amounts of data 
for their training~\cite{BenderEtAl2021}. 
In the case of so called \textit{chatbots} (conversational
agents intended for casual discussions on everyday topics) one
can source almost unlimited amounts of dialogue data from the
internet, e.g., twitter feeds. More often that not, however, such data
are of rather dubious quality and may contain significant
biases, involving language that is violent, racist, sexually explicit and so on~\cite{BenderEtAl2021}.
One of the supposed advantages of DNNs, being end-to-end systems, is their
ability to learn directly from data, without requiring any
human involvement (e.g., hand-coding) beyond preparing the data, the system architecture, and the training algorithm.
Yet that argument misses the extremely important point that the
data sets, which are generally huge, must be carefully curated when using
black box systems, as it is impossible to check whether they inadvertently
learn~\textit{unwanted} biases during training~\cite{BenderEtAl2021}.
It can thus be argued that the advantages of using such systems evaporate to a large 
extent, considering the massive amounts of work required to check the data.

A clear example of this is the case of Tay~\cite{Reese2016},
a chatbot released on the internet by Microsoft in 2016, which
was able to learn from its interaction with thousands of users. 
Because of a lack of quality control in the interactions, within
a few hours Tay developed nasty traits, expressing racist
and sexist views, forcing permanent shutdown of the agent.
Furthermore, it was recently shown that since DNNs can memorize part of the training data within their parameters, it is possible to carry out cyber-attacks to retrieve potentially privacy-sensitive information initially present in the data~\cite{CarliniEtAl2020}.

An even more serious issue appears when considering the use
of black box systems for so called \textit{task-oriented agents} that,
unlike chatbots, are intended to provide precise, consistent, and detailed
interaction with users regarding specific topics (e.g., in healthcare~\cite{Luxton2020}), thus requiring significant knowledge on those topics, beyond the mere general knowledge expected from a chatbot. 
A recent disastrous example concerns the use of GPT-3 as a conversational agent intended to provide assistance for people with mental health problems. Here, a user 
(thankfully only \textit{simulating} a patient suffering from depression) was \textit{advised} by the black box system to commit suicide~\cite{Daws2020}.
Thus, it is clear that black box systems cannot just be na\"ively adopted and expected to make no mistakes. In all fairness, it is often the case that the authors of black box models warn about these problems (see, e.g.,~\cite{BrownEtAl2020}).

Third, the prevailing (and rising) legal trends do \textit{not}
favor black box systems: There is proposed legislation
both in the US (the algorithmic accountability act) and in the
EU (the right to an explanation) that clearly favors more transparent
and accountable approaches than those offered by black box systems.
Incidentally, this is also a case where most of the ongoing research
is at odds with the desires and requirements of end users, for whom
transparency and accountability are clearly favored~\cite{JobinEtAl2019}.

One may thus wonder whether other research avenues exist, which are better suited for transparent and accountable conversational AI. 
In fact, preceding the advent of black box systems, many different approaches to dialogue
systems were considered, ranging from simple template-matching and finite-state systems~\cite{Wallace2009,JurafskyMartin2009}, to much more sophisticated (but also less transparent) approaches such as those based on partially observable Markov decision processes
(POMDPs)~\cite{YoungEtAl2013}. 
At the time, transparency and accountability were not as relevant and were not explicitly addressed.
This essentially translates to the fact that, today, simpler approaches enjoy better transparency mostly because they are in great part hand-crafted and remain somewhat limited, and not because they were explicitly designed to facilitate transparency and accountability.
Other modern approaches exist regarding language processing that are very different from DNN-based ones (see, e.g.,~\cite{CoeckeEtAl2020,Saba2007,Saba2018}), however these do not have transparency as a primary focus either.
We argue that one should strive to design systems that are \emph{both} transparent \emph{and} complex enough to enable human-like interactions. 
To that end, we believe that it is of primary importance to study what should be the design principles to ensure that transparency will be preserved as a conversational system grows increasingly complex.

\section{Design principles for transparent conversational AI}
We propose five key design principles that, in our view, should be incorporated
in conversational systems in order to ensure transparency, accountability, and safe use.
\subsection{Interpretability}
\label{sect:interpretability}
Contrary to the principles of black box systems, we
propose, as the first design principle, the use of
of interpretable primitives in conversational systems, i.e.~components that perform high-level operations such that both their purpose and mode of operation are easily human-understandable. Examples of such
components are program constructs such as sorting operations, fetch commands, comparison conditions, and so on, which can be combined to carry out very complex operations that still remain human-readable.

We also suggest following the so called \textit{pipeline model} (see e.g.~\cite{WahdeVirgolin2021})
whereby input processing, 
cognitive processing (decision-making), and output processing are clearly separated steps.
For the case of input processing, various options could be used, even black box models, as long as they are ultimately mapped
to a set of explicit intention sentences, which act as entry points for the cognitive processing.
In cases where the input does not perfectly
match any such intention sentence (which can happen, for example, with an approximation provided by a black box model), the agent should notify the user that it may require a clarification. 

The output processing (that follows the
cognitive processing) should also involve clearly defined, interpretable output sentences but, in this case, a semantic grammar or a black box model could be used for \textit{altering} the exact formulation of the output, without
changing its semantics, thus allowing a more life-like interaction.

The crucial middle step (cognitive processing), where the agent carries out deliberation and decision-making should, in our view, be
implemented in a generic fashion (exemplified below), using
a set of interpretable primitives that make it
easy to follow (and correct, when necessary) every step of the agent's decision-making as well as allowing re-use of implemented sequences of such primitives in other agents.
A simple example may help to illustrate this process. Consider an agent
that is capable to answer questions on geography and demographics, thus
containing a database with information about countries, cities, and so on. Now,
in this example, the user asks the agent \textit{which is the second largest
city in France?} After identifying the user's input (details omitted here),
the agent then carries out (in pseudo-code) the following generic cognitive processing \\
\texttt{FindAll(country=France, category=city)}$\rightarrow$\texttt{List1}, \\
\texttt{SortAscending(List1,population)}$\rightarrow$\texttt{List2}, \\
\texttt{GetElement(List2, 2)}$\rightarrow$\texttt{Element}, \\
\texttt{ExtractValue(Element, name)}$\rightarrow$\texttt{cityName}. \\
In plain English, the agent thus starts by finding all data items pertaining
to cities in France, then sorts the resulting list in ascending order
based on population, then extracts the second element from the list and, finally
extracts the \texttt{name} property from that element, storing it such that it can be accessed by the output processing step
that follows the cognitive step. Note that the sequence of primitives, in addition to being transparent, is completely generic: The
same sequence (but with other input parameters) could be used, say,
in an on-board information system on a bus, for answering questions
of the form \textit{What is the second stop after this?}

\subsection{Inherent capability to explain}
\label{sect:explainability}
Another key principle is that a conversational agent should be
able to \textit{explain} its reasoning, in a non-technical manner,
allowing both developers and end users to understand how the
agent reached a particular conclusion. This ability is particularly
important in cases where the agent's conclusion differs from what
the user expected. As an example of such a situation, one can
consider an agent that offers decision support to a driver
(of a truck, forklift, or any other vehicle). In a case where the
agent suggests a counter-intuitive maneuver, obtaining a clear explanation
may be crucial.

With the approach described in Subsection~\ref{sect:interpretability} above,
where the cognitive processing is generated as a sequence of elementary,
generic primitives, it is quite straightforward to include the ability to
explain, since every primitive of this kind can be associated with a
description of itself, taking into account also the input parameters.
Continuing on the example given in Subsection~\ref{sect:interpretability},
where the agent answers the question \textit{what is the second largest
city in France?}, the automatically generated explanation (from the agent) could read:
\textit{First, I found all data items pertaining to cities in France. Then, I sorted 
them in ascending order based on population. Next, I got  the second element 
from that list, and extracted its} \texttt{name} \textit{property}.

\subsection{Independent data}
We further propose that an agent's
memory should be divided into two clearly separated
parts: The \textit{procedural memory} that encodes the dialogue
capabilities described in the previous subsections, and
the \textit{declarative memory} that contains the facts
known to the agent.

Along with the generic cognitive
processing described above, this clear separation makes
it possible to replace parts (or all) of the declarative
memory, e.g.~application-specific parts, without having to change the procedural parts.
As an example, for an agent used as a museum guide, one
can easily just replace the declarative memory in order
to use the agent at a different museum, or even in
a different task, e.g., as a city guide or a tour guide.
Thus, with this approach, for many applications it is
possible to use an agent in an off-the-shelf manner, just providing it with a
suitable declarative memory.

\subsection{Interactive learning}
In the early days of conversational AI, agents were typically
generated using hand-coding, i.e.~a process whereby the developer
would implement, by hand, the dialogue capabilities of the agent.
With the advent of black box systems,
this process was largely replaced by an automated approach involving 
machine learning using large dialogue data sets for training.
The supposed advantage with the latter approach is that, in avoiding
hand-coding, one can generate better performing conversational agents, without
time-consuming manual implementation. However, as noted in
Section~\ref{sect:relatedwork}, the machine learning approach has its
own drawbacks, not least the need for careful checking of (vast amounts of) data,
to avoid unwanted biases.

Here, as the fourth key design principle, we argue that interactive learning whereby a person (acting as a tutor) teaches a machine new capabilities~\cite{PingEtAl2020,YinEtAl2017}, should be a primary approach when building conversational agents.
Coupled with the inquisitiveness (described below),
this method has the advantage of resembling the manner in which humans, not least children, 
learn new concepts. Just like hand-coding, this approach
also gives the developer full control over what
the agent learns but, unlike hand-coding, interactive learning makes
it possible also for a non-expert to teach an agent new capabilities.
Moreover, similarly
to what was described in Sect.~\ref{sect:explainability}, the interpretable primitives could be equipped with simple sentences by which they can be identified in a learning situation. A
specific example (cf.~the example in Sect.~\ref{sect:interpretability}) should serve to illustrate the principle: \\
User: \textit{I will teach you how to answer the question 
\lq\lq what is the $n^{\rm{th}}$ largest city in a country $x$\rq\rq}.\\
Agent: \textit{OK, please go ahead}. \\
User: \textit{First, find all city data items in $x$}. \\
Agent: \textit{Yes}. \\
User: \textit{Then sort the items in ascending order based on population}. \\
Agent: \textit{Understood.} \\
User: \textit{Next, find the $n^{\rm{th}}$ element}. \\
Agent: \textit{OK.}\\
User: \textit{Then extract and return the name property from that element.} \\
Agent: \textit{All right.} \\
At the end of this process, the agent will have assembled part of the cognitive processing as per the example of Sect.~\ref{sect:interpretability}.

\subsection{Inquisitiveness}
The final key design principle involves an agent's ability to actively
seek information and thus to expand its capabilities, for example by attempting to
make connections between different items in the data, categorizing data, and 
so on, in interaction with the user. Here, the agent should make \textit{hypotheses}, 
but not commit anything to its memory without first checking with the user, 
thus making sure that the agent will not accidentally learn things that it should not.
The inquisitiveness will also reduce the risk of accidentally
omitting important aspects during the interactive learning: The agent should
act as an active, eager student, rather than just passively receiving information. A simple example could be as follows: \\
Agent: \textit{I know that the planet Mars has two moons. I also know of another planet called Venus. Does it have moons?} \\
User: \textit{No it does not.} \\
Agent: \textit{OK. What about the Earth?} \\
User: \textit{It has one moon.} \\
Agent: \textit{Understood.} 

\section{Discussion} 
We have presented five design principles that we believe are key for the development of interpretable and safe conversational AI. While we hope that many readers will agree with
the proposed principles, and some may even find them very natural, we remark that the currently popular black box models are ill-suited to adhere to these principles.

An important objection that might be raised against the position taken in this paper is the
fact that black box approaches display remarkable performance, partly due to the fact
that such approaches are eminently suited for current hardware, e.g.,~GPUs~\cite{Hooker2020},
while it is far from clear whether systems based on the principles proposed here would be
able to reach similar performance, given current computers. However, as argued in~\cite{Rudin2019},
there is no fundamental reason to believe that one would have to sacrifice performance to achieve better interpretability; at the very least, 
we argue that this issue should be thoroughly investigated.

Interpretability is meant to make the inner workings of the system as explicit and transparent as possible to those involved in the design of the agent. By exposing interpretable primitives and their connections that together define the cognitive processing, an agent built in this manner makes it possible for the developers to get a clear overview of the system and thus to track, understand, and, if needed, correct its actions.
By contrast, the inherent capability to explain is meant to provide virtually \emph{any} user interacting with an agent with a clear explanation of the agent's reasoning.
The set of primitives need not be sufficiently expressive to carry out all forms
of computation, let alone achieve general AI; the primitives need only enable the conversational agent to carry out its task, while remaining sufficiently high-level so that even a combination of rather a large number of primitives would remain interpretable. The exact nature of the primitives (as well as the number of primitives required) is an important issue for further work and will
most likely depend on the application at hand.

The independent data principle encourages the design of conversational agents that are as general and decoupled from any particular application as possible, thus also making it possible to re-use existing cognitive capabilities across different application domains. 
Interactive learning and inquisitiveness are intended to provide a 
natural process for defining and improving conversational agents.
Systems built according to these principles provide an interface for defining, expanding, and correcting both the cognitive capabilities of the agent and its application-specific knowledge, making it possible for non-experts to seamlessly train a conversational agent.

There exists a number of  challenges for the development of conversational AI systems that adhere fully to these five principles. 
For example, it should be investigated what set of high-level primitives
would be required for the agent to be taught complex procedures via interactive learning.
Investing research efforts into interactive learning methods and inquisitiveness routines
is likely to be of crucial importance for developing systems that adhere to the
principles introduced here. Such methods need not themselves be interpretable (see, e.g.,~\cite{PingEtAl2020}), as
long as their final output, i.e.~the conversational agent, is interpretable.

Living in an age of big data, however, it would also make sense to try to
automatically infer some of the basic cognitive capabilities, in the form of combinations of the interpretable primitives, directly from data. 
We remark that, as long as interpretability is maintained, it will be possible (and necessary) to check and correct for the pick-up of unwanted biases when learning automatically from data prior to using the system in practice.
Note that, as soon as non-differentiable primitives are used (e.g., \texttt{IF-THEN-ELSE} conditions), one can no longer rely on gradient descent, but
instead will need alternative methods such as, e.g., automated program synthesis~\cite{ClarkeEtAl2003,Koza1992}.
Advancements in these fields may be needed before good conversational capabilities can be inferred directly from data.

Furthermore, we also do not exclude the use of black boxes, e.g., in the identification
of the user's intent. However, we are adamant in our belief that black box models should not be used in those parts of the agent that are directly concerned with \emph{decision-making}. If
the output of a black box component may have an influence on the decision-making (e.g., when used for identifying user intent), solutions should be studied to identify and minimize the related risks.

To conclude, in only the last few years it has become apparent that conversational AI is reaching a level of maturity such that its use will soon be widespread. Before long, in daily life one might encounter high-quality conversational agents whose output is almost indistinguishable from that of a person. At this crossroads, it is vital to consider the wider implications of conversational AI and to ascertain, to the greatest degree possible, that conversational agents are transparent and safe for use. To this end, we have here presented five key principles that, in our view, should be taken into account in research and applications involving conversational AI. We have also exemplified the application of those principles. In our current work, we are developing an interpretable approach that embodies all five principles, but it is important to note that the principles themselves are more important than any specific implementation. 

\bibliographystyle{./bibliography/IEEEtran}


\end{document}